\documentclass[12pt]{article}

\usepackage{times}
\usepackage{epsfig}
\usepackage{graphicx}
\usepackage{amsmath}
\usepackage{amssymb}

\begin{document}

\title{Simulating CRF with CNN for CNN}

\author{{Lena Gorelick} \and{Olga Veksler} }
\maketitle

\def\calP{{\cal P}}
\newcommand{\bx}{\mbox{\boldmath $x$}}
\newcommand{\vbar}{\:|\:}
\def\wpq{{w_{pq}}}

\begin{abstract}

Combining CNN with CRF for  modeling dependencies between pixel labels is a 
popular  research direction. This task is far from trivial, especially if
end-to-end training is desired. In this paper, we propose a novel simple approach to CNN+CRF combination. 
In particular, we propose to simulate a CRF regularizer  with a trainable module that has standard CNN architecture. 
We call this module a CRF Simulator. We can automatically generate an unlimited amount of ground truth for training such CRF Simulator without any user interaction,  provided we have an efficient algorithm for optimization of the actual CRF regularizer. 
 After our CRF Simulator is trained, it can be directly incorporated as
 part of any larger CNN architecture, enabling a seamless end-to-end training. In particular, the other modules can learn parameters that are more attuned to the performance of the 
 CRF Simulator module. 
We demonstrate the effectiveness of our approach on the task of salient object segmentation regularized with the standard binary CRF energy. 
In contrast to previous work we do not need to develop and implement the complex mechanics of optimizing a specific CRF as part of CNN.
In fact, our approach can be easily extended to other CRF energies, including multi-label. To the best of our knowledge we are the first to study the question of whether the output of CNNs can have regularization properties of CRFs. 
\end{abstract}

\section{Introduction}
\label{sec:intro}
The advances in Convolutional Neural Networks (CNNs)~\cite{Fukushima1980,LeCun:1989,Krizhevsky:NIPS2012,simonyan2014deep} lead to a tremendous success in computer vision
recently. CNNs excel at learning reliable features for a multitude of vision problems. Originally designed for classification of an entire image, CNNs have been extended to tasks where each image pixel is assigned a class label, such as semantic segmentation~\cite{farabet2013pami,FCN:Darrell},  stereo correspondence~\cite{Lecun:stereo_deep,DeepStereo:Chen}, 
optic flow~\cite{OpticFlow:Deep:2015}, etc.
 While the labels produced by CNN at nearby pixels
are correlated, due to the overlap of their receptive fields, such
dependencies  are not explicitly modeled. 
 Therefore, to improve the precision of pixel  labeling, CNNs are often combined with Conditional Random Fields (CRFs)~\cite{lafferty2001conditional,Koltun:DenseCRF}.  CRFs were designed for and excel at the task of modeling various pixel label dependencies.  CRF models, however, are not easy to incorporate with CNNs.
There are various prior  approaches.

The simplest approach to combine CNN and CRF is to apply CRF as  post processing~\cite{farabet2013pami,ChenPKMY14}. In these approaches, the probabilities for class labels learned by  CNN are converted into the unary CRF terms.  Then a CRF optimization algorithm, such as in~\cite{BVZ:PAMI01,Koltun:DenseCRF} is  used to obtain the final labeling. While simple, this approach 
does not support end-to-end training. The unary terms learned by CNN may not be the best tailored  to the separate CRF optimization step.

Another approach is to implement a CRF optimizer as part of the neural network. 
In~\cite{Zheng:CNN_CRF}, they show how to implement mean-field inference~\cite{Koller:2009:PGM}, a popular technique for optimizing full CRFs, as a Recurrent Neural Network (RNN).  This approach supports  end-to-end training, as CRF is integrated into the system architecture. The limitation is that the
CRF is fixed to the Gaussian-edge CRF from~\cite{Koltun:DenseCRF}, and the optimizer is
fixed to the mean-field inference. If another CRF and/or optimizer is desired, the architecture has to be hand-designed from scratch, which is technically
difficult.

Yet another  approach is to use the structured prediction learning framework~\cite{chen2015learning,DBLP:conf/cvpr/KnobelreiterRSP17}. Here the unary and/or pairwise potentials of a CRF are modeled with CNN-computed features. This approach supports end-to-end training. The difficulty here is optimization of the structured loss function. Computing the exact gradient of the loss is usually infeasible and has to be approximated. 
If CRF is restricted to be Gaussian, then inference is an easier task~\cite{Deep-GaussianCRF:2016,GaussianCRF:sparse:2016,chandra2017dense}.
However, the class of Gaussian CRFs  is rather limited. 

We propose a simple approach for combining CNNs with CRFs in one end-to-end trainable system. 
Our main idea is a special {\em CRF Simulator}  module. 
This module is built using standard CNN architecture and simulates the performance of a CRF regularizer. It is trained from data with ground truth obtained from an actual CRF optimizer. In particular, the input to CRF Simulator are the unary and pairwise terms of a CRF. The output is the simulation of a labeling that an actual CRF optimizer would produce.

 Note that our CRF Simulator module is trained to produce a labeling similar to that of an actual CRF optimizer, as opposed to training that aims to produce a low energy labeling.  Thus at no point during training we compute an actual CRF energy.  CRFs are used to produce a regularized labeling, for example a labeling with a short boundary, low boundary curvature, etc.  We can train a CRF-simulator that learns to produce a regularized labeling without an explicit knowledge of the energy function, just from observing the regularization effect on the training samples.  

While our approach requires labeled data, fortunately, a virtually unlimited amount of training data can be generated given an efficient actual CRF optimizer.
Once trained,  CRF Simulator is seamlessly incorporated as a part  of any larger CNN architecture. Our approach enables end-to-end training,
 namely the parameters of the other modules are learned with consideration of the CRF Simulator module behavior.

Another advantage of our approach is the ease of incorporating novel/different CRF
models. This is because our CRF Simulator is based on a standard CNN architecture. If we
wish to implement a new CRF regularizer, all we need to do is generate a new training dataset and re-train the CRF Simulator. This is possible when there is an efficient optimizer available for the CRF  of choice. Note that retraining a CRF simulator is significantly easier in practice,  compared to prior work,  where they have to develop a new technical approach for each new CRF model.

Regularizing and optimization~\cite{SZSVKATPAMI:2008,kappes_15_comparative}  is a large and well-established area of computer vision.
Multiple  regularizers and their properties were studied, as well as multiple optimization algorithms were developed over the past four decades.
It is an interesting question to ask if regularization properties can be
simulated by CNNs, without actually performing optimizaiton. As far as we know, our work is the first one to look at this question. 

We experimentally validate our proposed approach on the task of salient object segmentation.
This is a particularly suitable case since it requires optimization of binary CRF energy.  The ground truth data for training our  CRF simulator  is  obtained efficiently and exactly 
by a graph cut algorithm~\cite{BK:EMMCVPR01}.
Our approach can be easily extended to  the mutli-label case by  applying multi-label optimization algorithms to generate the ground truth for our simulator.
Fortunately, the area of efficient CRF optimization is vast and there are many options available,
see~\cite{SZSVKATPAMI:2008,kappes_15_comparative} for a review.

This paper is organized as follows. In Sec.~\ref{sec:related}, we review the related work. 
In Sec.~\ref{sec:crf_simulator}, we describe our CRF Simulator design. 
In Sec.~\ref{sec:complete_system} we show how to use  a CRF Simulator as a part of a complete 
CNN system. Experimental results are in Sec.~\ref{sec:experiments} and the conclusions and future work are
in Sec.~\ref{sec:conclusions}. 

\section{Related Work}
\label{sec:related}
The work in~\cite{farabet2013pami} was among the first to use a CRF for post-processing of
the results obtained with CNN. In particular, they use CNNs as a multi-scale feature extractor over superpixels.  
They form a superpixel-based CRF, and obtain the final segmentation using the expansion 
algorithm from~\cite{BVZ:PAMI01}.  
The work in~\cite{ChenPKMY14}  uses the Fully Convolutional Networks (FCN) 
from~\cite{FCN:Darrell} to learn the unary terms for a Gaussian edge 
full-CRF~\cite{Koltun:DenseCRF}.  Then mean field annealing is used for  CRF inference as a post-processing step. 
While simple, these approaches do not support end-to-end training. The CRF parameters are not learned from the training data together with CNN parameters. 

Another interesting approach that also uses Gaussian-edge full CRFs from~\cite{Koltun:DenseCRF} 
is in~\cite{Zheng:CNN_CRF}. They show how to implement mean-field inference~\cite{Koller:2009:PGM}, a popular technique for optimizing full CRFs, as a Recurrent Neural Network (RNN).  This approach is capable of end-to-end training, as CRF is integrated into the architecture of the whole system. Later, they extended their approach to handle higher order CRF 
terms~\cite{CRF_CNN_ECCV2016}. There are numerous other extensions to this RNN-CNN approach, 
such as  bilateral neural network for high dimensional sparse data~\cite{jampani:cvpr:2016}, more effective
CRF optimization based on linear programming LP~\cite{Kumar:DenseCRF_LP} or continuous relaxation~\cite{DBLP:journals/corr/DesmaisonBKTK16}.
Also see	~\cite{DBLP:journals/spm/ArnabZJRLKSRKT18} for a recent overview of the approaches that follow this direction. 

 While powerful, the RNN-CNN approach  requires hand-designing special architecture for each new CRFl one wishes to model. This is a very difficult task, and thus far, other than CRF model with mean field 
optimization, there are few other examples. The only other example we are aware of is
the work of~\cite{NIPS2017_6702}, that describe how to implement a  
submodular CRF layer within a CNN architecture. Their approach is technically difficult and is limited to 
submodular functions. 

In~\cite{chen2015learning}, they model MRF potentials as deep features using a structured learning framework. In this framework, computing exact gradient descent updates is computationally infeasible, and they resort to various approximations, such as using local belief functions instead of the true marginals. 
While theoretically interesting and amenable to  end-to-end training, their approach relies on a number of approximations and is complex to implement.

In~\cite{DBLP:conf/cvpr/KnobelreiterRSP17}, they also proposed a method for CRF-CNN training based on structured learning. Like~\cite{chen2015learning}, the main difficulty of this approach is efficiently back-propagating through the CRF module. For each specific CRF model, a new approach has to be developed from scratch. 

An alternative approach is to incorporate regularization directly into the loss function. For example the approach in~\cite{tang2018normalized} incorporates normalized cut regularization into the loss function for the problem of weakly supervised learning. However, incorporating length regularization into loss during fully supervised learning suprisingly leads to inferior performance compared with unregularized loss function\footnote{Private communication with the authors}, also confirmed by our experiments.

\section{CRF Simulator}
\label{sec:crf_simulator}
In this section we describe the design of our CRF Simulator. 
Even though we present the case of binary CRF energy for the task of salient object
segmentation, our proposed approach is general and can be easilty extended to other energies and
applications. 
In Sec.~\ref{sec:prelim}  we describe the CRF energy function that we seek to simulate. 
In Sec.~\ref{sec:architecture} we give details of the CNN architecture for our CRF Simulator. 
In Sec.~\ref{sec:training_data} we explain how we automatically generate the training data.

\subsection{CRF Energy Function}
\label{sec:prelim}
The  energy function in this section is the one proposed in~\cite{Boykov:IJCV2006} for 
segmenting an object from the background.
Let $\calP$ be the set of image pixels, and $x_p \in \{0,1\}$ be the label  assigned to pixel $p$. Here 
label $0$ corresponds to the background, and $1$ to the salient object. 
Let  $\mathbf{x}= (x_p\vbar p\in \calP)$ be a vector storing  labels  for all pixels.  The CRF energy that 
we wish to simulate is defined as
\begin{equation} 
f(\mathbf{x})=\sum_{p\in  \calP} f_p(x_p) + \sum_{p,q\in \calP}  \wpq \cdot  [x_p\neq x_q],
\label{eq:main-energy}
\end{equation}
The unary terms $f_p(x_p)$ give the cost of assigning pixel $p$  to label $x_p$. We model them
as negative log probabilities for salient object and background. 
The pairwise terms $\wpq \cdot  [x_p\neq x_q]$ penalize label discontinuities between neighboring pixels. We set
$\wpq$ to be inversely proportional to the color difference of pixels
$p$ and $q$
\begin{equation}
\label{eq:wpqInside}
\wpq = \lambda \cdot  exp \left (  -\frac{||C_p-C_q||^2}{2  \sigma^2 } \right ).
\end{equation}
 This type of CRF energy function regularizes
a labeling by encouraging a shorter boundary length. The larger is the $\lambda$, the stronger is the
 regularizer influence in comparison to the unary terms. 

The optimum labeling minimizng the energy in Eq.~\eqref{eq:main-energy} can be found via the minimum cut
algorithm~\cite{BK:EMMCVPR01}. Thus this energy function is particularly suitable for an
initial evaluation of our approach since we can generate  globally optimum labelings 
for our ground truth dataset. 
 
\subsection{CNN Architecture}
\label{sec:architecture}
\begin{figure*}
\begin{center}
\includegraphics[width = 0.96\textwidth]{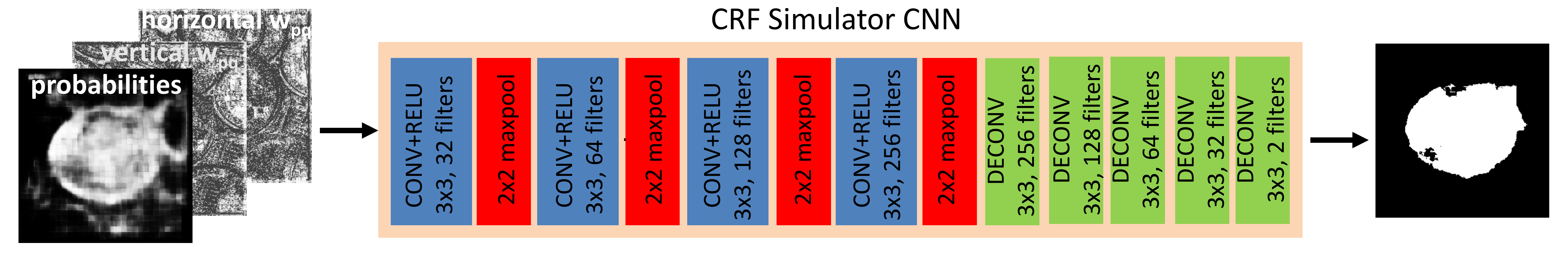} 
\end{center}
\caption{Architecture of CRF Simulator CNN.  
 \label{fig:CRF-CNN}} 
\end{figure*}

To train our CRF Simulator, we use the network illustrated in 
Fig.~\ref{fig:CRF-CNN}. 
The input is  3-channel ``image'' where the first channel stores the unary terms of the energy function in  Eq.~\eqref{eq:main-energy} and the other two channels store horizontal and verticall pairwise terms respectively.The output is the regularized labeling. The architecture is a standard Encoder-Decoder one, with four convolutional and deconvolutional layers. 
This simple network achieves  $90.44\%$ F-measure on the test data. We also tried to get a better performance by using features from the VGG network~\cite{simonyan2014deep} pretrained on ImageNet~\cite{imagenet_cvpr09}. However, such network had a slightly inferior performance. The likely explaination is that VGG network is pretrained on real world images, whereas the input to CRF Simulator CNN is of different type.

\subsection{Training Data}
\label{sec:training_data}
The input to our CRF Simulator is the unary and pairwise terms 
of the energy in Eq.~\eqref{eq:main-energy}. First we describe the input unary terms.
For training our CRF-simulator, we wish to have a richer class of optimization problems than just those coming from salience. Therefore we  combine two data sets: the MSRA-B dataset~\cite{Learning:Salient:TPAMI2010} and PASCAL data set 2012~\cite{pascal-voc-2012}. We divide MSRA-B data set into 2500 images for training, 500 for validation, and 2000 for testing. Similarly, we take the first 80\% images from PASCAL database for training, the next 10\% for validation and the last 10\% for testing giving subsets of  (13700, 1712, 1712) images respectively. 

We then obtain input unary terms using two different procedures, one for images from MSRA-B data set,  where the data terms are based on pixel salience, and one for images from PASCAL data set, where the data terms are based on appearance.

\begin{figure*}
\begin{center}
\includegraphics[width =1\textwidth]{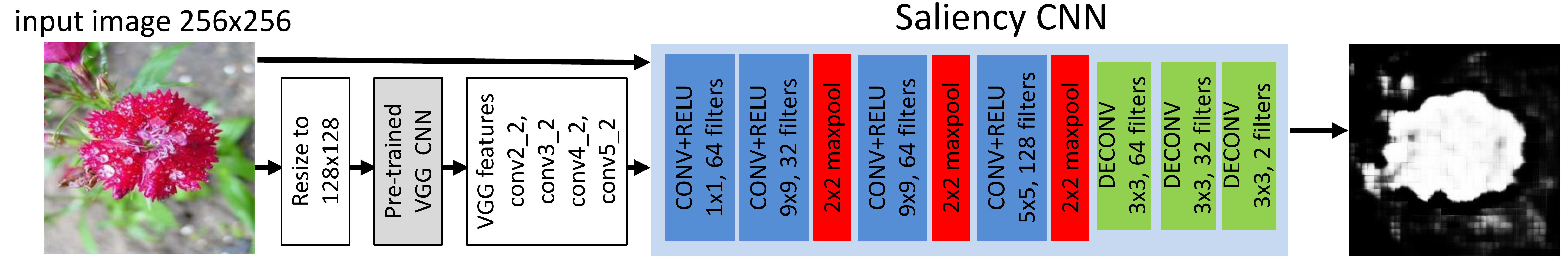}
\end{center}
\caption{Architecture of Saliency CNN. 
 \label{fig:saliency-CNN}} 
\end{figure*}

In the first case we use the training subset of MSRA-B to train a separate Saliency-CNN component shown in Fig.~\ref{fig:saliency-CNN}. For each input image it outputs salience probability map. This probability map is subsequently used as an input to the CRF Simulator. See  Sec.~\ref{sec:saliency-CNN} for detailed description of Saliency-CNN.  

In the second case, for each PASCAL image we create multiple instances of input data-terms. For any image $i$ that has an object segmentation map  provided with the data set, we compute data-terms $\mathbf{f}_i^k(\mathbf{x})$ separately for each object mask $k$ as follows. 
First, we fit an appearance model for foreground and background of each mask $k$ 
using discrete normalized color histograms with 16 bins per channel. We then normalize the resulting data terms to sum to one for each pixel.
For images in PASCAL that do not have object segmentation masks we fit histograms to four random rectangular regions. 
This process results in several instances of data terms $\mathbf{f}_i^k$ per image, yielding additional 
(50068, 6852, 6852) input images for training, validation and testing of  our CRF Simulator respectively.

During training, we use a set of 30  different $\lambda$ values, ranging from $0$ to $400$. These values are sampled more densely  in the lower range, since lower values of $\lambda$ lead to a better segmentation more frequently. 
When $\lambda = 400$, for all images in our dataset  the CRF-optimizer produces an empty labeling.
Thus  $\lambda = 400$  is a sufficiently large value to ensure CRF Simulator learns the extreme case when the object is completely smoothed out by the regularizer.

The training examples for our CRF Simulator are ``images'' with three channels. The
first channel stores the  unary terms as described above: either from Saliency CNN in Fig.~\ref{fig:saliency-CNN} or from fitting appearnce models to segmentation masks in the PASCAL dataset. The second and third channels are the horizontal and vertical terms $\wpq$  respectively, vectorized as two images. 
energy in Eq.~\eqref{eq:main-energy} completely. 
Tto generate the ground truth for each training example, we simply run the 
the min-cut/max-flow algorithm of~\cite{BK:EMMCVPR01} producing the labeling of the lowest energy.
We use training, validation and test subsets from MSRA-B dataset, and 30 different values of $\lambda$ to generate
75,000 training, 15,000 validation and 60,000 test examples for our CRF Simulator. Similarly, for images from  PASCAL dataset we run max-flow to produce labeling for 16 different values of $\lambda$ and randomly choose five of them to generate additional 250340, 34260 and 34260 examples for training, validation and testing respectively.

\section{Complete System}
\label{sec:complete_system}

\begin{figure}
\begin{center}
\includegraphics[width = 0.5\textwidth]{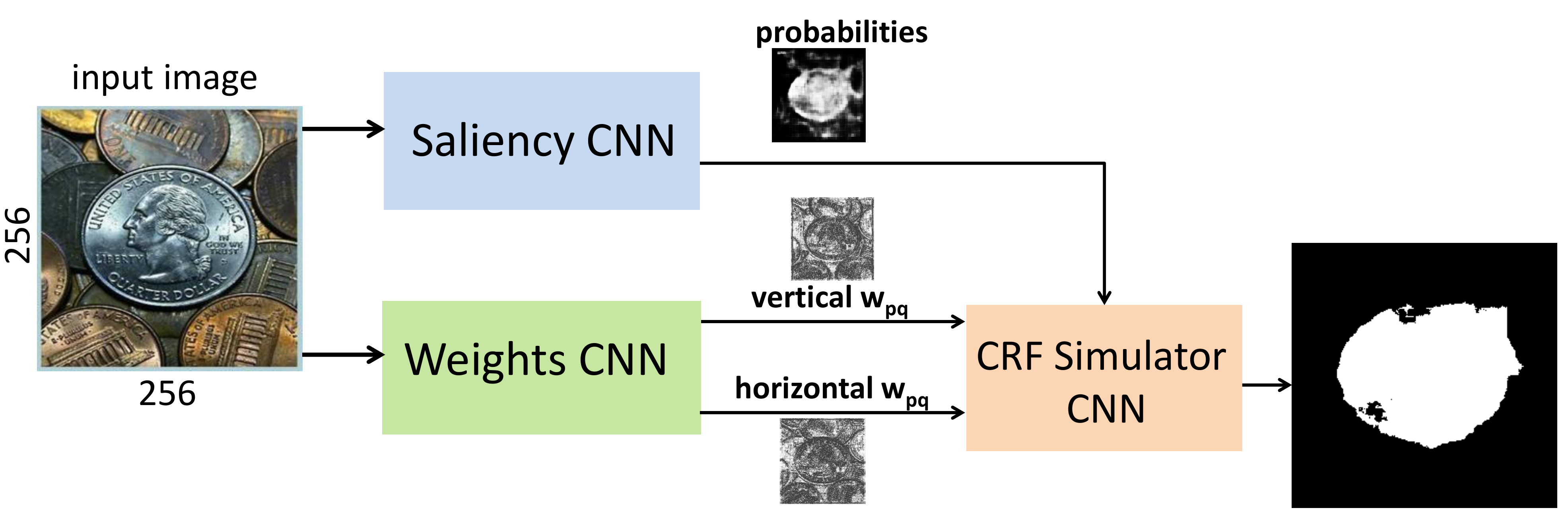}
\end{center}
\caption{Complete system overview.  
 \label{fig:complete_system}} 
\end{figure}
We now describe how we use our CRF Simulator in a complete system for segmenting salient objects, see Fig.~\ref{fig:complete_system}. Our system has three main components: Saliency-CNN, Weights-CNN and CRF Simulator CNN. Components are illustrated in detail in Fig.~\ref{fig:saliency-CNN}, Fig.~\ref{fig:weights-CNN} and Fig.~\ref{fig:CRF-CNN} respectively.  The input image first goes through Saliency-CNN which outputs a single channel of salience probabilitiy map. The original image is also  used as input to the Weights-CNN component which outputs two channels of vertical and horizontal smoothness terms  $\wpq$. Then, the salience probabilities and the smoothness terms are concatenated and fed as input to the pre-trained CRF Simulator CNN which outputs regularized segment of the salient object.

Before training the complete system we initialize  
CRF Simulator module with the weights obtained during the separate
training stage on the CRF Simulator dataset described in Sec.~\ref{sec:training_data}. 
Similarly, Saliency CNN weights are initialized 
with the weights obtained during separate training on the MSRA-B saliency dataset. The parameters of the Weights CNN module are initialized randomly. During training, we can choose to either keep the weights of CRF Simulator and Saliency CNN fixed, or allow them to be further tuned. We evaluate these options 
in Sec.~\ref{sec:experiments}.

\subsection{Saliency-CNN}\label{sec:saliency-CNN}
Given an input image, the role of the Saliency-CNN is to output a salience map.
Our architecture for this module is in Fig.~\ref{fig:saliency-CNN}. 
We make use of the VGG network~\cite{simonyan2014deep} pre-trained on ImageNet~\cite{imagenet_cvpr09}. The input image is resized to 128x128 and fed into VGG CNN. We then use VGG features in a manner similar to \cite{li2018interactive}. Namely,  we extract conv2\_2, conv3\_2, conv4\_2 and conv5\_2, upscale them to the original size and concatenate to the original image. Similarly to \cite{li2018interactive} we reduce the dimensionality  by performing one layer of 1x1 convolution which results in 64 features. We then use a standard Encoder-Decoder network, with four convolutional and deconvolutional layers as shown in Fig.~\ref{fig:saliency-CNN}. The output is a salience probability map.
Our network achieves F-measure of $88.21\%$  on the MSRA-B dataset.   
The state of the art performance is close to $F$-measure of $93\%$~\cite{hou2017deeply}. Our network is relatively close to the state of the art, and offers a good compromize in terms of its size. 

\subsection{Weights-CNN}\label{sec:weights-CNN}
\begin{figure}
\begin{center}
\includegraphics[width = 1\columnwidth]{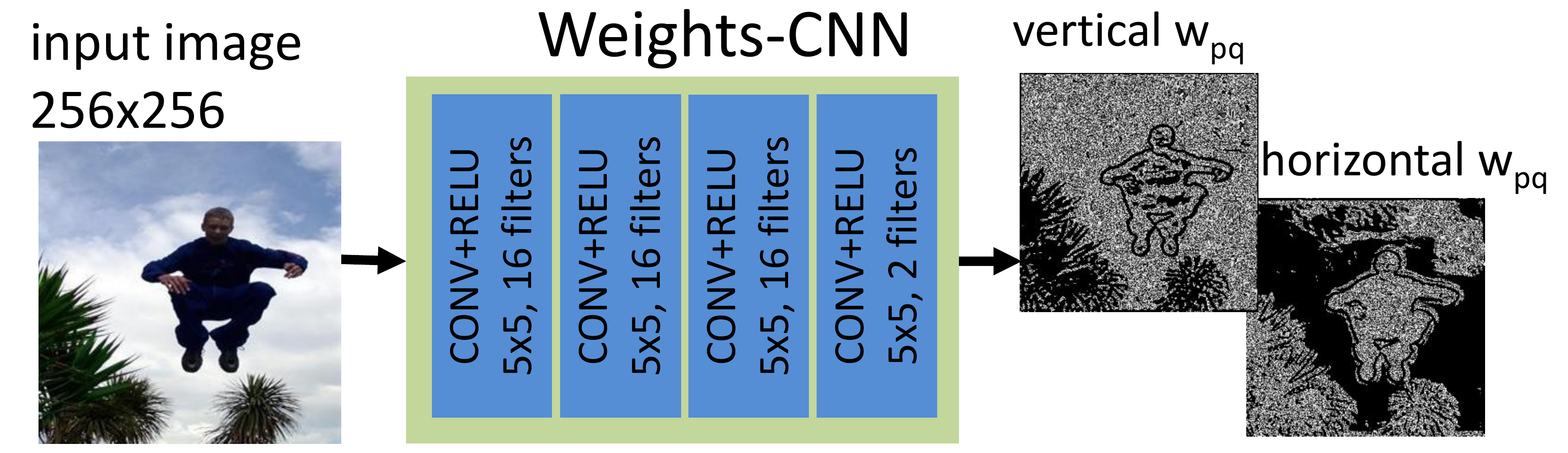}
\end{center}
\caption{Module for obtaining $\wpq$ for the complete system.
 \label{fig:weights-CNN}} 
\end{figure}

The idea of Weights-CNN component is that the smoothness terms and the strength of regularization useful for CRF Simulator can be learned directly from the image. 
Our architecture for this component is in 
 Fig.~\ref{fig:weights-CNN}. It takes the input image and performs four sets of 5x5 convolutions. The last layer outputs two channels. The first one is taken to be the vertical $\wpq$, and the last channel the horizontal $\wpq$. Although there is no explicit $\lambda$ parameter, the strength of regularization is implicitly learned  through the relative magnitude of weights $\wpq$: the larger are $\wpq$, the more regularization CRF Simulator will  produce.

\section{Experimental Results}
\label{sec:experiments}
We used the MSRA-B saliency dataset from~\cite{Learning:Salient:TPAMI2010} to evaluate our complete system.
We divided it randomly into 2500 images for training, 500 for validation, and 2000 for testing.  We re-scaled all images to be of size $256$ by $256$.

We use the $F$-measure as the main criteria for performance evaluation, since this is the customary measure for  salient object segmentation.  Let $P$ be the precision (i.e.  
the percentage of pixels that are correctly labeled as object out of all pixels labeled as object), and $R$ the recall (i.e. the percentage of pixels that are correctly labeled as object out of all object pixels). Then $F$-measure is defined as
\begin{equation}
F_{\beta} = (1+\beta^2) \frac{P\cdot R}{\beta^2 \cdot P +R }.
\end{equation}
We set $\beta^2 = 0.3$.

The dataset for  training CRF Simulator contains some cases where the ground truth has no object pixels. This happens when $\lambda$ value is too high. In these cases, the $F$-measure is not well defined. To fix this problem, we reverse the meaning of object and background when computing the $F$-measure for such examples. Intuitively, this makes sense since for such examples, the
most accurate answer  is to label every single pixel as ``background''. 

All CNNs were trained with cross-entropy loss, Adam optimizer~\cite{Adam:2014} and  learning rate of $10^{-4}$ for
100 epochs.

\subsection{CRF-simulator vs. real CRF-optimizer}
\label{sec:crf_simulate_real}
We first evaluate the performance of our CRF Simulator.
We report the accuracy with which CRF Simulator simulates the output of an actual CRF optimizer and analyze the energy of the labelings that CRF Simulator produces. 
We use  the dataset described in Sec.~\ref{sec:training_data}. 

\subsubsection{Accuracy}
\label{sec:crf_sim_accuracy}
\begin{figure*}
\begin{center}
\includegraphics[width =1.0\textwidth]{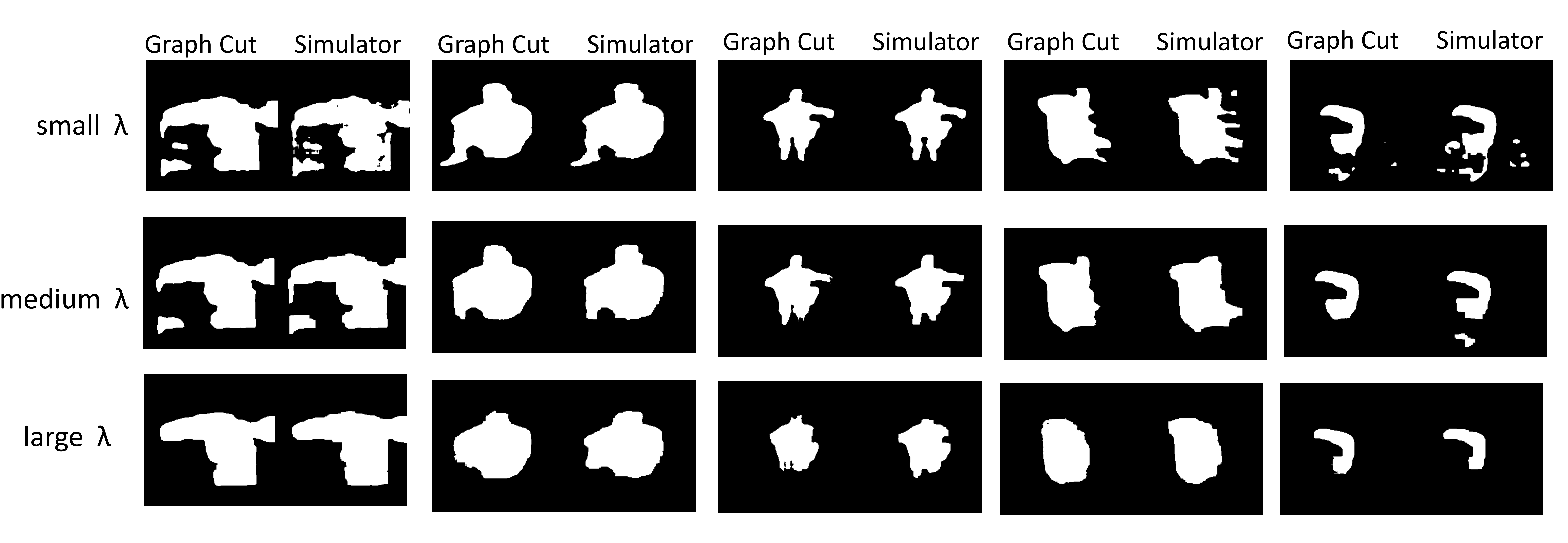}
\end{center}
\caption{Illustrates some labellings learned by our CRF Simulator for small, medium, and large $\lambda$ values. 
In each image, left part shows output of graph cut, and right part shows output of our simulator. 
 \label{fig:LearnCut}} 
\end{figure*}

Our CRF Simulator CNN achieves  $90.44\%$ F-measure on the test data. 
Several examples obtained by  CRF-simulator are shown in Fig.~\ref{fig:LearnCut}, along with the ground truth segmentations produced by the actuall CRF optimizer.
We selected examples for a small, medium, and large values of   $\lambda$.  The labelings produced by our simulator are not guaranteed to have the lowest energy, since energy is not a part of training criteria. However, they capture the ``spirit''
of the regularization encoded into the energy in Eq.~\eqref{eq:main-energy}. For low $\lambda$ values, the obtained labeling is noisy, as mostly unary terms are followed. For larger $\lambda$, 
noise and thin structures are  smoothed out, holes are filled in, etc. One can argue that it is the spirit of the regularizer that one is after
when performing energy optimization, and the actual energy value  is of no interest by itself.

 It is interesting to note 
that for $\lambda = 0$, our labelings do not follow the unary terms exactly, unlike the exact CRF optimizer.
Our simulator still performs a small amount of regularization. 
This is, again, because our simulator learns to mimic the behavior of the actual optimizer instead of learning to produce the lowest energy labeling. 

\subsubsection{Energy}
\label{sec:crf_cnn_eng}
\begin{figure*}
\begin{center}
\begin{tabular}{ll}
\includegraphics[width =0.4\textwidth]{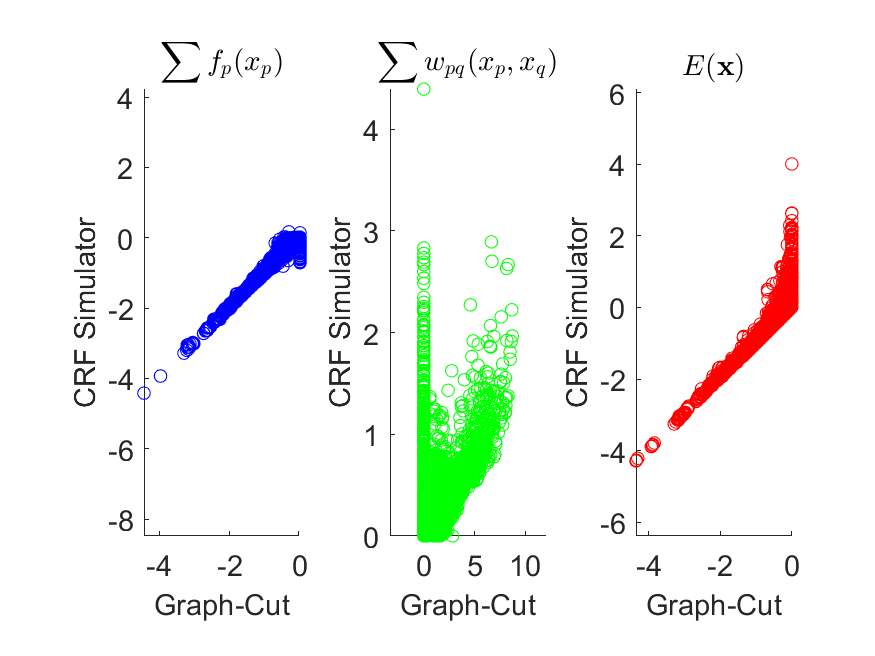} &
\includegraphics[width =0.4\textwidth]{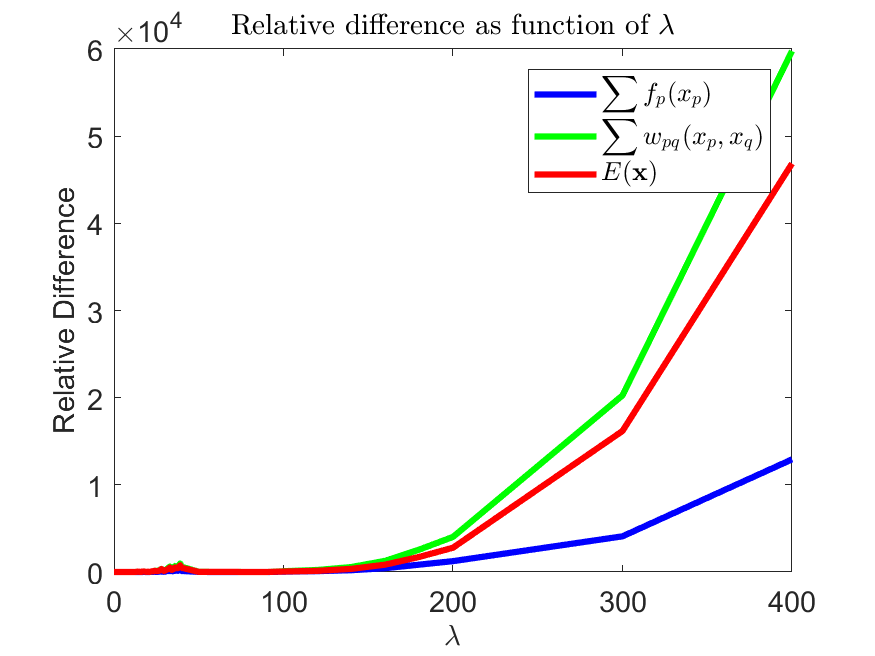}
\end{tabular}
\end{center}
\caption{Left: Scatter plots illustrating correlation between the optimal energy and the energy obtained with CRF-simulator on the test images. Unary terms (with correlation coefficient $r=0.98$) are shown in blue, smoothness terms ($r=0.61$) in green and total energy ($r=0.93$) in red. Right: Average relative difference in energy as a function of regularization weight $\lambda$.
 \label{fig:scatter_plots}} 
\end{figure*}

It is interesting to analyze the energy of the labelings produced by our CRF Simulator and compare to the energy of the optimal labelings produced by an actual CRF regularizer.
Figure~\ref{fig:scatter_plots}(left) shows scatter plots where each test image is a point, x-axis is the optimal energy obtained with graph-cut and y-axis is the energy computed for the labeling obtained with CRF-simulator. The plot in blue compares just the unary terms of the energy showing correlation coefficient of $r=0.98$. The plot in green plots just the smoothness terms of the energy showing correlation coefficient of $r=0.61$. Finally, the plot in red plots the total energy showing correlation coefficient of $r=0.93$. Observe the region of the green scatter plot directly above graph-cut energy 0. This region dramatically reduces the correlation coefficient. It consists of images for which optimal solution is empty and therefore has zero smoothness term. This happens for high values of regularization weight $\lambda$. Insuch case, the learned solution is usually non-empty and has a non-zero discontinuity cost, reducing correlation. We further analyzie the relative difference bewteen the optimal and simulated energies as a function of regularization weight $\lambda$, see Fig.~\ref{fig:scatter_plots}(right). Indeed, the relative difference growth as a function of $\lambda$.  In practice however, we are interested in the lower range of values for the regularization weight $\lambda$ where correlation of energies is high.

\subsection{Complete System}
\label{sec:complete_system}
In this section, we evaluate the performance of the complete end-to-end system  that includes 
CRF-simulator component (Fig.~\ref{fig:complete_system}). But first, for completeness, we report the performance of the
Saliency-CNN alone (without post-processing) and with actual CRF optimizer postprocessing.

\paragraph{Saliency CNN:}
Our Saliency CNN achieves the F-measure of $88.21\%$ on the task of saliency segmentation, when used on its own. 
\paragraph{Saliency CNN followed by CRF optimizer postprocessing:}
Next we apply actual CRF optimizer as a postprocessing step. The results depend on the choice of $\lambda$ in the energy function, which controls how much regularization is needed. In general, the
the optimal value of $\lambda$ varies between the images. However, when applying CRF Optimizer as post-processing, we must fix $\lambda$ to a certain value.  Choosing the best fixed value per dataset (using validation data), $\lambda=15$ gives the best performance, with the overall $F$-measure of $88.81\%$. 
A complete end-to-end trained system that uses our CRF Simulator 
(see Sec.~\ref{sec:complete_system}) achieves a better
$F$-measure of  $89.40\%$

For CRF-optimizer, it is also interesting to measure what is the best $F$-measure that can be achieved if we choose
the best $\lambda$ for each image. While unrealistic in practice (best $\lambda$ per image is not known), this gives an idea of
how much improvemnt a regularizer can achieve if it was possible to learn best weights per each individual image. 
Going over all $\lambda$'s for each image, and choosing the one that gives the best $F$-measure for that image,
gives the overall performance of $91.22\%$.  Thus learning better weights would help the complete system performance. 

\begin{table*}
\begin{center}
\renewcommand\arraystretch{1.5}
\begin{tabular}{|c |c|c| c|c| c| c|}
 \hline
Saliency CNN &    CRF optimizer & Complete  TF &  Complete  TT &  Complete  FT & Complete FF& Complete Random \\
\hline
\hline
$ 88.22$ &  $ 88.81 $  & $89.3$ & $\bf{89.40}$ & 89.05 &  $ 88.93 $  &   $88.35$ \\
 \hline
\end{tabular}
\vspace{1ex}
 \caption{$F$-measure on the test data for  the Saliency CNN, Saliency CNN with CRF Optimizer as 
post-processing,  and our five Complete systems.  }\label{table:summary}
\end{center}
\end{table*}

\begin{figure*}
\begin{center}
\includegraphics[width = 0.7\textwidth]{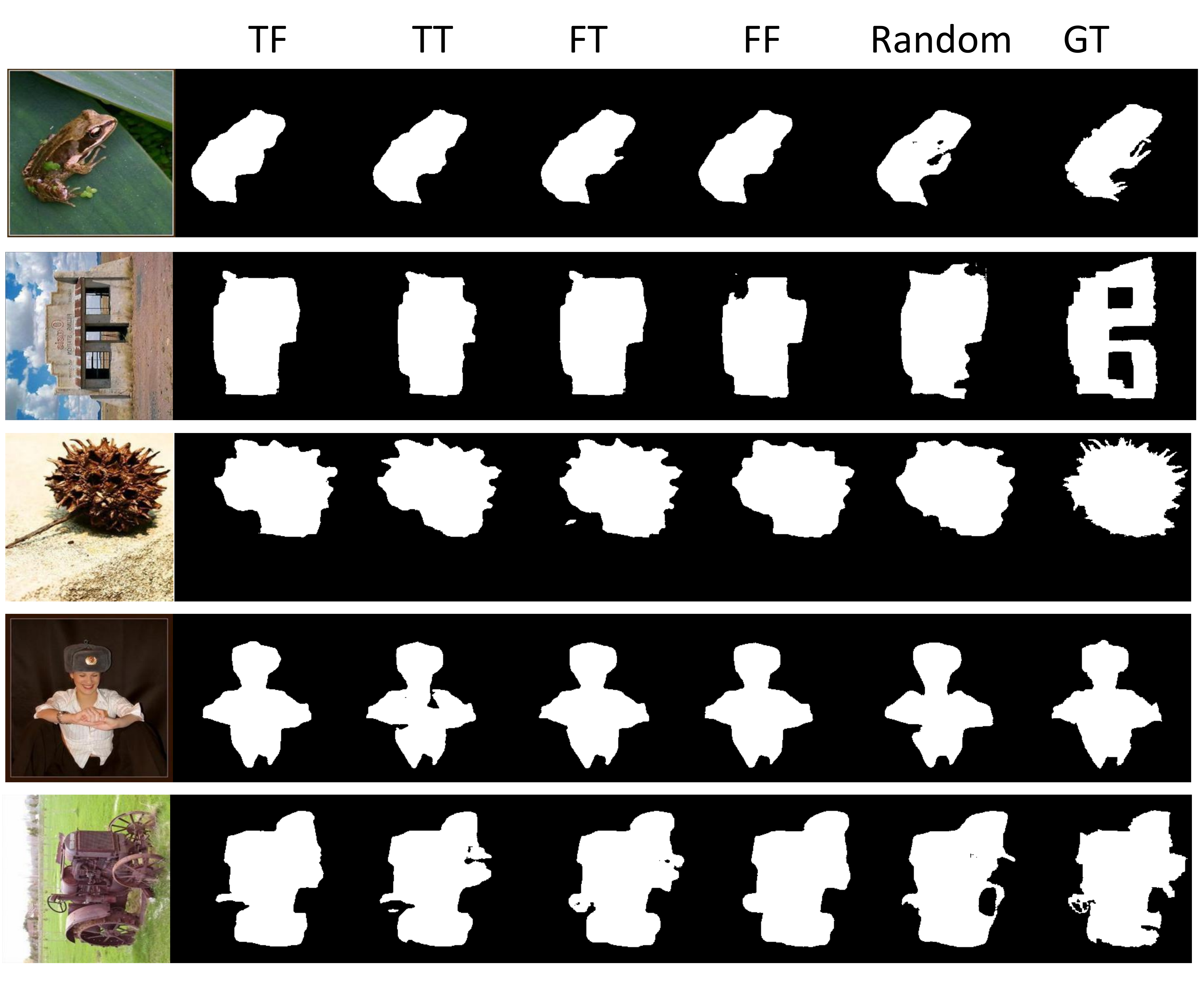}
\end{center}
\caption{
Some representative results of our Complete system.\label{fig:examples}} 
\end{figure*}

\begin{figure}
\begin{center}
\includegraphics[width = 0.45\textwidth]{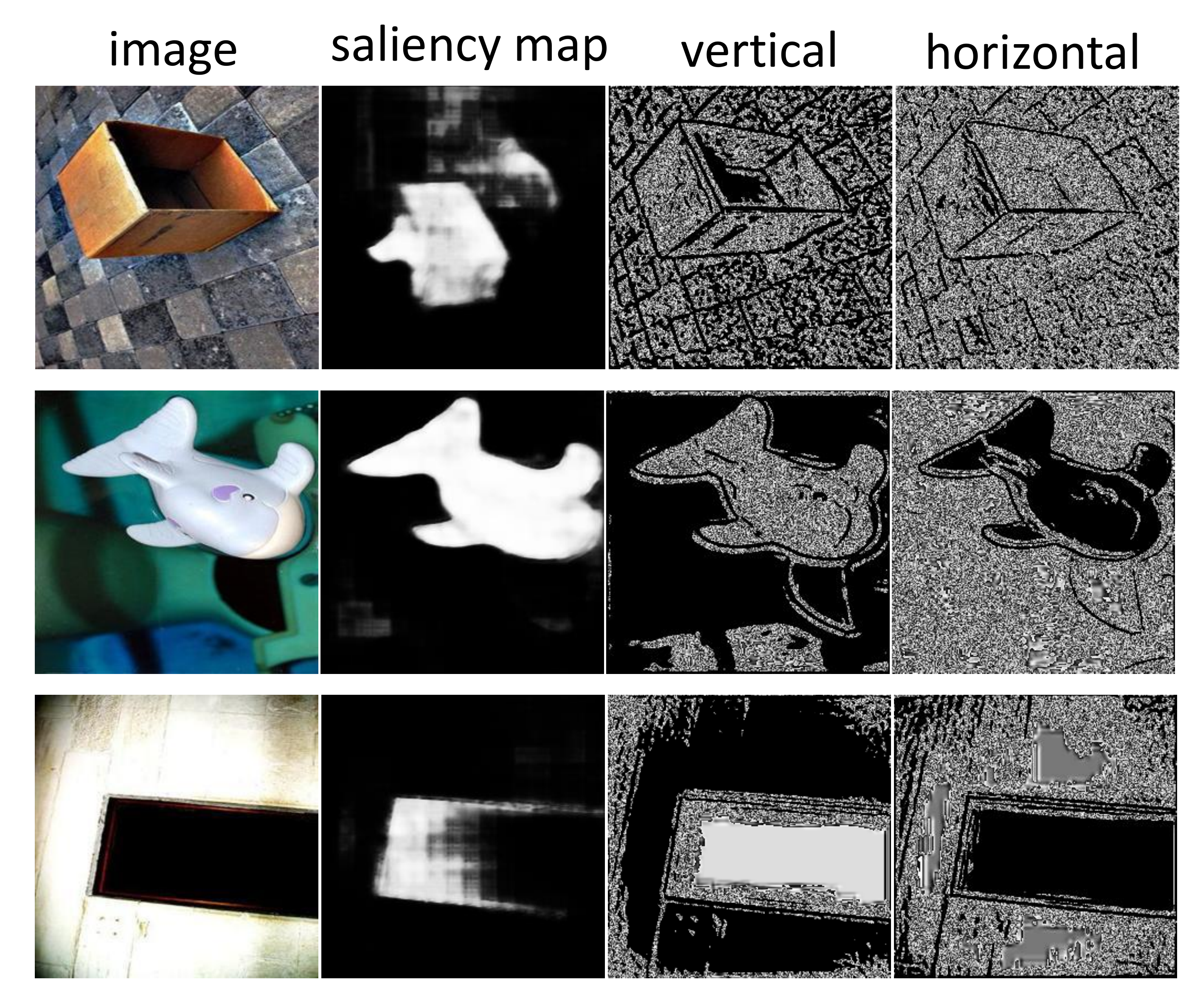}
\end{center}
\caption{Edges learned by the complete system. From left to right: input image, saliency probabilities, vertical and horizontal $\wpq$.
 \label{fig:edges}} 
\end{figure}

\paragraph{Complete System Variants:}
We train our Complete System in five different regimes, depending on which CNN components have 
their  weights fixed or allowed to be tuned during training. 
The parameters of our Weights CNN component are always trainable.  The weights of Saliency CNN and CRF-Simulator
CNN can either be fixed or trainable.  This gives us four different combinations during training. The fifth regime is as follows. We allow all components to have trainable weights but initialize CRF Simulator weights to be random. 
 This allows us to validate that the gains we have with CRF Simulator component are not just due to having a larger network. 

Table~\ref{table:summary} shows the $F$-measure on the test data
for the Saliency CNN,  Saliency CNN with CRF optimizer as post-processing, 
and our Complete System in five different regimes.
Two boolean flags after ``Complete'' specify which weights of the Complete system are fixed and which are allowed to be tuned. The first boolean flag corresponds to the Saliency CNN, and the second to the CRF Simulator. For example, ``TF'' means the Saliency CNN weights are tuned, while CRF Simulator weights are fixed.  Finally, the last column shows the peformance for our complete system when all the weights are tuned, but CRF Simulator  weights are initialized to be random.

CRF Optimizer, used as post-processing, improves saliency results. The Complete system, initialized with random weights (last column in Table~\ref{table:summary} ) is better than Saliency CNN alone, but  actually  worse than CRF Optimizer post-processing.
This is not surprizing since in this case, CRF Simulator module does not perform any regularization but only serves to enarge a saliency network.  All four variants of our Complete System with pre-trained CRF Simulator perform better than CRF Optimizer as post-processing.  The best performance is in the ``TT'' case, since there are more tunable weights. 
The ``TF'' case is better than ``FT'' case. It is better to leave CRF Simulator weights unchanged but tune saliency weights as opposed to leaving Saliency weights fixed and tuning CRF Simulator weights. In the ``TF'' case, the complete sytem has a chance to fine-tune the Saliency module weights to suite better to the regularization transformations done by the CRF Simulator module.

Some labelings obtained  by our Complete System are shown in Fig.~\ref{fig:examples}.   Observe that labelings produced by Complete System variants with fixed CRF Simulator (second and fifth column) capture the spirit of the CRF regularizer better. They even imitate the rectilinear nature of boundaries characteristics of  four neighborhood CRF optimizer. Also observe that labelings in the ``Random'' column are the least regularized.

It is interesting to inspect weights $\wpq$ learned by our Weight CNN module in
Fig.~\ref{fig:weights-CNN}.
Consider some examples in Fig.~\ref{fig:edges}. Here each row shows the input image, the  saliency map, and
the vertical and horizontal weights $\wpq$ (brighter intensity means larger weight). 
For most training examples the learned horizontal and vertical weights $\wpq$ resemble edges as shown in the top row of Fig.~\ref{fig:edges}. 
For about a third of the data, the learned weights are similar to those shown in the second row of Fig.~\ref{fig:edges}. Large portions of the background (for the vertical terms) and large portions inside the salient object (for horizontal terms) are of low cost with almost no edges detected. This might be because these images are particularly easy for saliency detection. It is possible that while Weights CNN module was designed to learn smoothness terms, it also learns about saliency to some degree. Therefore it does not have to detect edges in areas where it is easy to determine whether the pixelsa belong to salient object or background. In a few very rare cases, the edges
look like the bottom row in  Fig.~\ref{fig:edges}. This may be either failure to learn edges well. Or it may have something to do with the failure of the 
Saliency CNN (Fig.~\ref{fig:saliency-CNN})
to obtain a good saliency map. Notice in this case, the saliency map is missing a large portion of the object. The edge learning component may have imposed a large penalty on vertical weights to try to prevent  the CRF Simulator cutting through the salient object.

\section{Conclusions and Future Work}
\label{sec:conclusions}
We presented a new approach for combining CNN and CRF in an end-to-end trainable system.
The main idea is to simulate the regularization properties of an actual CRF optimizer by a special
 module. This module has a standard CNN architecture and is trained on examples produced
by an actual CRF optimizer. The ground truth can be obtained in an unlimited quantity without user interaction, given
an efficient CRF optimizer. The advantage of our approach over previous work is 
that we do not have to implement the complex mechanics of optimizing CRF as part of CNN.

In the future, we plan to investigate CRF-simulators for other energy/optimizer combinations. In particular, an interesting question is how well CRF-simulator performs for energies where the optimum solution cannot be computed exactly, and has to be approximated.

Another interesting direction is to design a better Weight CNN module for learning
regularization weights. There is ample evidence that learning to adapt regularizer strength appropriately for each image can lead to  significant performance gains.

{\small
\bibliographystyle{ieee}
\bibliography{cvpr2019}
}

\end{document}